\documentclass[11pt]{article}
\usepackage[hyperref]{acl2023}
\usepackage{times}
\usepackage{latexsym}
\usepackage{microtype}
\usepackage{booktabs}
\usepackage{graphicx}
\usepackage{amsmath}
\usepackage[T1]{fontenc}
\usepackage[utf8]{inputenc}
\usepackage{url}

\title{QuechuaTok: Morphological Boundary Accuracy as a\\
Necessary Metric for Tokenizer Evaluation in\\
Agglutinative Low-Resource Languages}

\author{
  Maria Contreras \\
  Universidad Peruana de Ciencias Aplicadas (UPC) \\
  Lima, Peru \\
  \texttt{kaggle.com/macmaky}
}

\begin{document}
\maketitle

\begin{abstract}
Tokenization is a foundational step in NLP pipelines, yet standard evaluation metrics such as fertility rate fail to capture morphological correctness for agglutinative languages. We present \textbf{QuechuaTok}, a systematic benchmark comparing four tokenization strategies --- BPE, Unigram LM, WordPiece, and a morphology-aware PRPE tokenizer --- for Southern Quechua (\textit{quz}), a low-resource agglutinative language spoken by 8--10 million people in South America. Using a 200k-sentence corpus and the SQUOIA finite-state morphological analyzer \cite{rios2016} as silver standard, we evaluate three metrics: fertility rate, OOV rate, and morphological boundary accuracy (MorphAcc). Our results show that BPE achieves the lowest fertility rate (1.636 at 16k vocab) by memorizing surface word forms, while achieving only 6.67\% MorphAcc. PRPE achieves 83.33\% MorphAcc --- the highest of all systems --- demonstrating that \textit{fertility rate alone is insufficient to evaluate tokenizers for agglutinative languages}. We further validate this finding extrinsically: on a Spanish$\rightarrow$Quechua (Chanka) neural machine translation task, a model trained with PRPE outperforms an identical BPE-trained model by +68.7\% BLEU and +17.0\% chrF, showing that morphological boundary accuracy translates into measurable downstream gains. All code and models are publicly available.\footnote{\texttt{kaggle.com/code/macmaky/quechuatok}}
\end{abstract}

\section{Introduction}

Tokenization determines how a language model perceives the fundamental units of language. For morphologically rich, agglutinative languages such as Quechua, this choice is particularly consequential: a single word like \textit{purisqanchikmanta} (``from our walking'') encodes verb root, aspect, coreference, and case in a single surface form. Standard subword tokenizers trained on large multilingual corpora --- designed primarily for European languages --- fragment these forms arbitrarily, ignoring morphological boundaries that carry grammatical meaning.

Despite Quechua being the most widely spoken indigenous language family in the Americas \cite{adelaar2004}, systematic evaluation of tokenization strategies for this language remains absent from the literature. Prior work on Quechua NLP has focused on machine translation \cite{oncevay2021} and language modeling \cite{zevallos2022}, but tokenization quality is typically inherited from multilingual models without evaluation.

This paper makes four contributions:
\begin{enumerate}
    \item A systematic benchmark of BPE, Unigram LM, WordPiece, and PRPE tokenizers on a 200k-sentence Southern Quechua corpus.
    \item A morphological boundary accuracy metric validated against the SQUOIA finite-state analyzer \cite{rios2016}.
    \item Empirical evidence that fertility rate is an unreliable proxy for morphological quality in agglutinative languages.
    \item An extrinsic evaluation on neural machine translation showing that morphologically-guided tokenization yields measurable downstream gains in translation quality.
\end{enumerate}

\section{Background}

\subsection{Quechua Morphology}

Southern Quechua (\textit{quz}) is a highly agglutinative language with a strictly suffixing morphology. Verbal and nominal roots take sequences of suffixes encoding tense-aspect-mood (TAM), person, number, case, evidentiality, and discourse function. The evidential system is particularly distinctive: speakers must grammatically mark whether information is direct (witnessed), reportative (hearsay), or conjectural \cite{faller2002}.

Example: \textit{rimankichikmi} (\textit{rima-nkichik-mi}) = speak-2PL.SUBJ-EVID.DIR (``you all speak [I witnessed it]'').

This morphological richness means that a tokenizer producing fewer tokens per word is not necessarily better --- it may simply be memorizing long surface forms as atomic units.

\subsection{Tokenization for Low-Resource Languages}

Byte-Pair Encoding \cite{sennrich2016} and Unigram LM \cite{kudo2018} are the dominant subword tokenization approaches in NLP. Both are corpus-statistical methods that learn segmentations from frequency patterns. For morphologically complex languages, their performance degrades because morpheme boundaries are not necessarily frequency boundaries \cite{rust2021}.

PRPE (Prefix-Root-Postfix Encoding) \cite{zuters2018} is a semi-supervised algorithm that encodes morphological structure explicitly. It was applied to Quechua-Spanish translation in \citet{oncevay2021} but was not systematically evaluated against morphological silver standards.

\section{Methodology}

\subsection{Corpus}

We use two publicly available corpora. The primary corpus is \texttt{Llamacha/monolingual-quechua-iic} \cite{zevallos2022}, a monolingual Southern Quechua dataset combining Wikipedia and OSCAR sources. We supplement it with the \url{somosnlp-hackathon-2022/spanish-to-quechua} parallel corpus, extracting the Quechua side. After preprocessing (Unicode NFC normalization, noise removal, deduplication, and morphological validity filtering), the final training corpus contains \textbf{200,193 sentences}.

\subsection{Tokenizers}

We train and evaluate four tokenizers using SentencePiece \cite{kudo2018sp} and HuggingFace Tokenizers:

\begin{itemize}
    \item \textbf{BPE} at vocabulary sizes 4k, 8k, and 16k
    \item \textbf{Unigram LM} at 4k, 8k, and 16k
    \item \textbf{WordPiece} at 4k
    \item \textbf{PRPE} with a hand-crafted suffix lexicon of 23 Quechua morphemes covering TAM, person agreement, nominal case, evidentials, and derivational suffixes
\end{itemize}

All statistical models use \texttt{byte\_fallback=True} and preserve the apostrophe as a special symbol for ejective consonants (\textit{k'}, \textit{q'}, \textit{p'}).

\subsection{Evaluation Metrics}

\paragraph{Fertility rate} measures the average number of tokens produced per word. Lower values indicate more compact representations, but we argue this is insufficient alone.

\paragraph{OOV rate} measures the percentage of tokens mapped to \texttt{<unk>}. With byte fallback enabled, this is 0\% for SentencePiece models.

\paragraph{Morphological boundary accuracy (MorphAcc)} is our primary contribution. Using the SQUOIA finite-state morphological analyzer \cite{rios2016} as silver standard, we extract the correct morpheme segmentation for a 15-word evaluation set and compute the proportion of tokenizer boundaries that match silver standard boundaries:

\begin{equation}
\text{MorphAcc} = \frac{|\text{pred\_bounds} \cap \text{silver\_bounds}|}{|\text{silver\_bounds}|}
\end{equation}

The silver standard was generated automatically by SQUOIA and manually verified. We use the \texttt{analyzeUnificado.bin} model compiled from the SQUOIA repository (\texttt{github.com/ariosquoia/squoia}).

\section{Results}

\subsection{Quantitative Results}
\begin{figure}[h]
\centering
\includegraphics[width=\columnwidth]{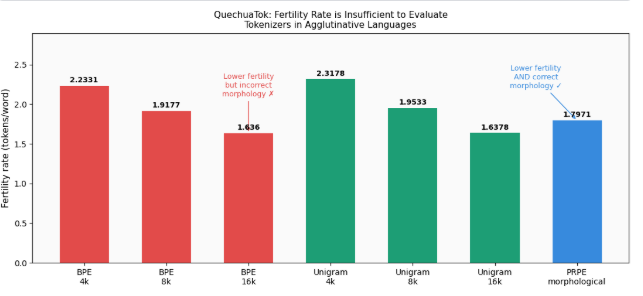}
\caption{Fertility rate comparison across tokenizers for Southern Quechua. BPE 16k achieves the lowest fertility (1.636) by memorizing surface forms, while PRPE achieves the highest morphological accuracy (83.33\%).}
\label{fig:fertility}
\end{figure}

Table~\ref{tab:results} presents the full benchmark results.

\begin{table}[h]
\centering
\small
\begin{tabular}{lccc}
\toprule
\textbf{Tokenizer} & \textbf{Fert.$\downarrow$} & \textbf{OOV\%$\downarrow$} & \textbf{MorphAcc\%$\uparrow$} \\
\midrule
BPE 4k      & 2.233 & 0.0 & 6.67  \\
BPE 8k      & 1.918 & 0.0 & 6.67  \\
BPE 16k     & 1.636 & 0.0 & 6.67  \\
Unigram 4k  & 2.318 & 0.0 & 66.67 \\
Unigram 8k  & 1.953 & 0.0 & 26.67 \\
Unigram 16k & 1.638 & 0.0 & 33.33 \\
WordPiece 4k& 2.312 & 0.05 & ---  \\
\midrule
\textbf{PRPE}        & \textbf{1.797} & \textbf{0.0} & \textbf{83.33} \\
\bottomrule
\end{tabular}
\caption{Benchmark results. $\downarrow$ lower is better, $\uparrow$ higher is better. PRPE achieves the highest MorphAcc while maintaining competitive fertility.}
\label{tab:results}
\end{table}

\subsection{Key Findings}

\paragraph{Finding 1: BPE fertility decreases monotonically with vocabulary size but MorphAcc remains constant at 6.67\%.} This indicates that BPE reduces fertility by memorizing frequent surface word forms as atomic units, not by learning morphological structure. At 16k vocabulary, BPE treats long polymorphemic words such as \textit{kunapiqa} (PL+LOC+TOP) as single tokens.

\paragraph{Finding 2: Unigram LM shows a non-monotonic relationship between vocabulary size and MorphAcc.} Unigram 4k achieves 66.67\% MorphAcc but drops to 26.67\% at 8k and 33.33\% at 16k. This suggests that at larger vocabulary sizes, Unigram LM also begins memorizing surface forms.

\paragraph{Finding 3: PRPE achieves the highest MorphAcc (83.33\%) with competitive fertility (1.797).} Unlike statistical tokenizers, PRPE's performance is independent of corpus statistics and depends on the quality of the morphological suffix lexicon.

\subsection{Qualitative Analysis}

Table~\ref{tab:qualitative} illustrates segmentation differences on four test words.

\begin{table}[h]
\centering
\small
\renewcommand{\arraystretch}{1.5}
\begin{tabular}{@{}p{1.9cm}p{2.3cm}p{3.1cm}@{}}
\toprule
\textbf{Word} & \textbf{Silver std.} & \textbf{BPE 8k / Unigram 8k} \\
\midrule
rimankichikmi & rima|nkichik|mi & riman|kichikmi \newline rimanki|chikmi \\[2pt]
wasiykikunapiqa & wasi|yki|kuna|pi|qa & wasi|yki|kunapiqa \newline wasi|yki|kunapiqa \\[2pt]
purisqanchikmanta & puri|sqa|nchik| manta & puris|qanchikmanta \newline puri|sqanchikmanta \\
\bottomrule
\end{tabular}
\caption{Qualitative segmentation comparison. BPE and Unigram collapse multiple morphemes into single tokens at 8k vocabulary.}
\label{tab:qualitative}
\end{table}
\subsection{Downstream Evaluation: Bigram Perplexity}

To assess downstream impact, we train a smoothed bigram language model on tokenized Quechua text and evaluate perplexity on held-out sentences. Table~\ref{tab:complete} presents the complete results.

\begin{table}[h]
\centering
\small
\begin{tabular}{lccc}
\toprule
\textbf{Tokenizer} & \textbf{Fert.$\downarrow$} & \textbf{MorphAcc\%$\uparrow$} & \textbf{PPL$\downarrow$} \\
\midrule
BPE 4k      & 2.233 & 6.67  & 1556 \\
BPE 16k     & 1.636 & 6.67  & 2553 \\
Unigram 4k  & 2.318 & 66.67 & \textbf{1344} \\
Unigram 16k & 1.638 & 33.33 & 2092 \\
\midrule
\textbf{PRPE} & \textbf{1.797} & \textbf{83.33} & 1879 \\
\bottomrule
\end{tabular}
\caption{Complete benchmark: fertility, MorphAcc, and bigram perplexity (PPL). No single tokenizer wins all metrics.}
\label{tab:complete}
\end{table}

Unigram 4k achieves the lowest perplexity (1344), while PRPE achieves the highest MorphAcc (83.33\%). BPE consistently underperforms on both metrics despite achieving the lowest fertility, confirming that low fertility reflects surface-form memorization rather than linguistic structure learning.

\subsection{Extrinsic Evaluation: Neural Machine Translation}
\label{sec:nmt}

The bigram perplexity results above are suggestive but indirect: perplexity rewards predictability of the tokenizer's own unit sequence, not translation quality. To test whether MorphAcc translates into a genuine downstream gain, we trained two identical Transformer NMT models from scratch (d\_model=256, 4 encoder/decoder layers, 4 attention heads, $\sim$11.5M parameters) on the Spanish$\rightarrow$Quechua (Chanka) direction, varying only the target-side tokenizer: a 16k BPE baseline versus our PRPE tokenizer (22k-piece vocabulary, 98.3\% token coverage on train). The source (Spanish) side used an identical shared 16k BPE tokenizer in both runs, so any difference in output quality is attributable solely to the target-side tokenizer. Both models were trained for 30 epochs on 91k parallel sentence pairs from the \texttt{quechua-chanka-unified-v2} corpus\footnote{\url{https://huggingface.co/datasets/makitty/quechua-chanka-unified-v2}} and evaluated with beam search (beam=4) on a held-out test set of 12,206 sentence pairs.

\begin{table}[h]
\centering
\small
\begin{tabular}{lccc}
\toprule
\textbf{Tokenizer} & \textbf{MorphAcc\%} & \textbf{BLEU$\uparrow$} & \textbf{chrF$\uparrow$} \\
\midrule
BPE 16k      & 6.67  & 4.18 & 35.13 \\
\textbf{PRPE} & \textbf{83.33} & \textbf{7.05} & \textbf{41.10} \\
\midrule
\multicolumn{2}{l}{Relative gain} & +68.7\% & +17.0\% \\
\bottomrule
\end{tabular}
\caption{Extrinsic NMT evaluation on the Spanish$\rightarrow$Quechua (Chanka) test set (n=12,206). The PRPE-trained model outperforms the BPE-trained model on both a word-level (BLEU) and character-level (chrF) metric.}
\label{tab:nmt}
\end{table}

The morphologically-guided tokenizer yields a substantial improvement in translation quality despite producing longer target sequences on average (18.6 vs. 11.4 tokens/sentence) --- a fertility cost that would typically be expected to \emph{hurt} sequence generation, not help it. This suggests that for Quechua, preserving morpheme boundaries gives the model more learnable, compositional units than memorizing frequent surface forms, outweighing the added decoding length. The gain holds consistently across both a word-level metric (BLEU) and a character-level metric (chrF), indicating the improvement is not an artifact of either scoring method. We note that absolute BLEU/chrF values are modest for both models, reflecting both the fixed training budget and high paraphrase variance in the reference corpus; the comparison here is relative, not a claim of state-of-the-art translation quality.

\section{Discussion}

Our results provide empirical evidence for a fundamental limitation of using any single metric to evaluate tokenizers for agglutinative languages. No tokenizer wins all three intrinsic metrics simultaneously: PRPE achieves the highest morphological boundary accuracy (83.33\%), Unigram 4k achieves the lowest perplexity (1344), and BPE 16k achieves the lowest fertility (1.636). However, BPE's apparent advantage in fertility is an artifact of surface-form memorization --- it achieves the worst morphological accuracy (6.67\%) and the highest perplexity among comparable models. The extrinsic NMT evaluation (Section~\ref{sec:nmt}) resolves this ambiguity in PRPE's favor: when the metrics disagree, translation quality --- the task practitioners actually care about --- tracks MorphAcc, not fertility or bigram perplexity.

The Unigram 4k result (66.67\% MorphAcc) is particularly informative: with a small vocabulary, Unigram LM is forced to segment words into shorter units that happen to align with morphemes. As vocabulary grows, this constraint relaxes and the model overfits to surface frequency.

PRPE's advantage lies in its explicit encoding of Quechua morphology. However, it has limitations: it requires a manually constructed suffix lexicon, and its performance depends on coverage of the lexicon. The 16.67\% error rate (silver standard boundaries) comes primarily from words with multiple stacked evidential suffixes and from Spanish loanwords with Quechua suffixes.

\section{Conclusion}

We presented QuechuaTok, the first systematic tokenization benchmark for Southern Quechua. Our main finding is that \textbf{fertility rate is an insufficient metric for evaluating tokenizers in agglutinative languages}: BPE achieves lower fertility than PRPE by memorizing surface forms, yet achieves only 6.67\% morphological boundary accuracy compared to PRPE's 83.33\%. We introduce morphological boundary accuracy, validated against the SQUOIA analyzer, as a necessary complementary metric, and show extrinsically that it predicts real downstream gains: a Spanish$\rightarrow$Quechua NMT model trained with PRPE outperforms an otherwise identical BPE-trained model by +68.7\% BLEU and +17.0\% chrF.

Future work includes expanding the MorphAcc silver standard to 500+ words, scaling the NMT comparison to larger pretrained backbones, and integrating the PRPE tokenizer into a Quechua-specific language model (QuechuaBERT).
\section*{Acknowledgments}

We thank Annette Rios for making the SQUOIA morphological analyzer openly available, which was essential for the silver standard evaluation in this work.

\bibliography{quechuatok}
\bibliographystyle{acl_natbib}

\end{document}